# Deep LG-Track: An Enhanced Localization-Confidence-Guided Multi-Object Tracker

Ting Meng, Chunyun Fu*, Xiangyan Yan, Zheng Liang, Pan Ji, Jianwen Wang, Tao Huang

*Abstract*—Multi-object tracking plays a crucial role in various applications, such as autonomous driving and security surveillance. This study introduces Deep LG-Track, a novel multi-object tracker that incorporates three key enhancements to improve the tracking accuracy and robustness. First, an adaptive Kalman filter is developed to dynamically update the covariance of measurement noise based on detection confidence and trajectory disappearance. Second, a novel cost matrix is formulated to adaptively fuse motion and appearance information, leveraging localization confidence and detection confidence as weighting factors. Third, a dynamic appearance feature updating strategy is introduced, adjusting the relative weighting of historical and current appearance features based on appearance clarity and localization accuracy. Comprehensive evaluations on the MOT17 and MOT20 datasets demonstrate that the proposed Deep LG-Track consistently outperforms state-of-the-art trackers across multiple performance metrics, highlighting its effectiveness in multi-object tracking tasks.

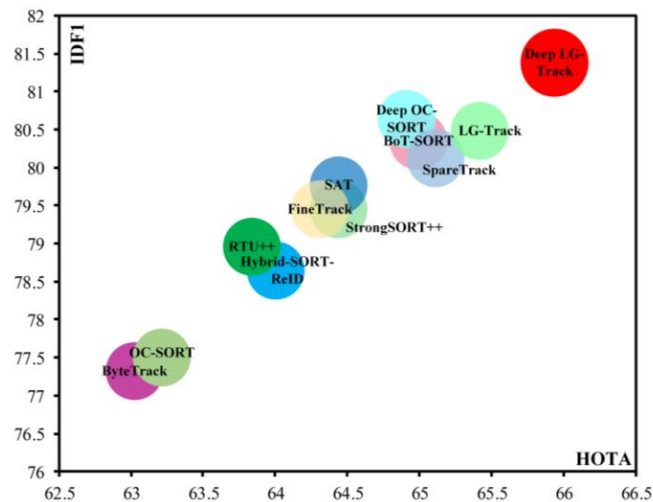

**Fig. 1.** Kalman Filter Performance comparison between Deep LG-Track and the SOTA trackers on the MOT17 dataset (IDF1-HOTA). The horizontal axis represents HOTA, and the vertical axis represents IDF1.

## I. INTRODUCTION

MULTI-Object Tracking (MOT) plays an essential role in the computer vision domain, which finds its application in various fields such as autonomous driving, safety monitoring, and robotics. The most widely used paradigm of MOT is tracking-by-detection (TBD), in which the MOT task is decomposed into two sequential stages – object detection and data association [1],[2].

The TBD-based MOT solutions currently face several constraints, notably in three aspects – trajectory prediction and update, data association (i.e. trajectory-detection matching), as well as appearance feature update.

First, in current TBD-based MOT solutions, a conventional Kalman filter is commonly employed for each trajectory to predict its location in the new frame and then update its status [3], [4], [5], [6]. Such solutions apply uniform measurement noise covariance across all objects without accounting for variations in detection quality. To mitigate this issue, various Kalman filter variants, such as the Noise Scale Adaptive Kalman Filter (NSAKF) [7] and the Power-Adaptive (PAKF) [8] have been proposed. Note that in Kalman filtering, prediction errors are bound to accumulate and the predicted boxes gradually deviate from the ground truth, if prediction is executed for multiple consecutive frames without state update. To handle the adverse effect of accumulated errors, deteriorated prediction results should be assigned less weight in the Kalman filtering process. However, this important issue has not been taken into account in the above methods.

Second, in the data association stage, the integration of motion information and appearance information of objects within a cost matrix has been a common approach in previous works [4], [9], [10], [11]. However, these methods simply employ experimentally determined, fixed-value factors as relative weights between the motion cost and the appearance cost. Such a fusion pattern faces challenges when encountering objects with different detection qualities. For example, some detection boxes may present clear appearance but inaccurate localization, while others may exhibit accurate location but unclear appearance. Apparently, in the cost matrix design, higher relative weighting should be assigned to the

This work was supported by the Shanxi Province Major Science and Technology Project under Grant 202301150401011.

Ting Meng, Chunyun Fu, Xiangyan Yan, Zheng Liang and Pan Ji are with the College of Mechanical and Vehicle Engineering, Chongqing University, Chongqing 400044, China (e-mail: mengting@stu.cqu.edu.cn; fuchunyun@cqu.edu.cn; yanxiangyan@stu.cqu.edu.cn; liangzheng@stu.cqu.edu.cn; jipan@stu.cqu.edu.cn).

Jianwen Wang is with the State Key Laboratory of Intelligent Vehicle Safety Technology, Chongqing Changan Automobile Co., Ltd., Chongqing 400023, China (e-mail: wangjw3@changan.com.cn).

Tao Huang is with the College of Science and Engineering, James Cook University, Smithfield, Queensland 4878, Australia (email: tao.huang1@jcu.edu.au).

appearance cost if the detection box is clear, and vice versa. Taking this issue into account, a mechanism should be incorporated in the cost matrix to adaptively adjust the relative weighting between motion and appearance costs, thereby reflecting the effects of detection quality on data association.

Third, the Exponential Moving Average (EMA) method has been traditionally employed for updating the appearance features of trajectories, as seen in studies [4], [9], [10], where a fixed weighting factor is used to fuse historical and current appearance features. This fixed fusion pattern does not consider detection quality, and as a result, appearance features of low-quality detections can lead to tracking performance deterioration. Intuitively, if the quality of current detection is low, a higher weight should be assigned to historical features, and vice versa. To mitigate this limitation, Maggiolino et al. [12] recently introduced Deep OC-SORT, which adopts a Dynamic Appearance (DA) strategy to dynamically adjust the weighting factor for appearance features based on detection confidence. It has been emphasized in LG-Track [1] the critical need to evaluate the detection quality of a detection box from two distinct aspects – localization accuracy and appearance clarity, rather than using the overall detection confidence as the only detection quality indicator. Apparently, the above DA strategy still relies on the overall detection confidence and does not consider localization accuracy and appearance clarity separately.

To address the above-identified limitations in current TBD-based MOT solutions, we introduce Deep LG-Track in this paper, an evolution of our previous multi-object tracker – LG-Track [1]. Deep LG-Track advances the performance of our original LG-Track through three major enhancements and demonstrates superior performance on the MOT17 and MOT20 datasets. Fig. 1 displays a performance comparison between Deep LG-Track and other leading methods on the MOT17 dataset.

Based on our original LG-Track, three core enhancements have been achieved in this study, as follows:

- **Adaptive Covariance of Measurement Noise**: We introduce an adaptive covariance matrix of measurement noise to reflect the variation of measurement uncertainty in Kalman filtering. An adaptive factor, dependent on detection confidence and trajectory disappearance, is designed to dynamically adjust the measurement noise.
- **Adaptive Cost Matrix**: An adaptive cost matrix is designed to optimally assign weights to the motion cost and the appearance cost. Each detection box's localization confidence and detection confidence are employed as weights in the cost matrix for fusing these two costs.
- **Strong Dynamic Appearance**: A refined appearance feature updating mechanism is introduced to dynamically adjust the relative weighting between historical and current appearance features. For a newly matched detection box, this relative weighting is made dependent on the detection box's appearance clarity and localization accuracy.

The above three core enhancements significantly improve the accuracy and robustness of data association, leading to superior performance of our proposed tracker. Extensive tests on the MOT17 and MOT20 datasets confirm that the proposed Deep LG-Track outperforms other state-of-the-art methods in terms of key performance metrics such as HOTA, AssA, and IDSW. These results solidly validate its effectiveness in MOT applications and innovative contributions to this domain.

The rest of this paper is organized as follows. Section II provides a brief review of related works in this field. Section III elaborates on our proposed Deep LG-Track, concentrating on the above-mentioned three core enhancements. Section IV provides details of our experimental verification, with both numerical and graphical tracking results. Section V concludes the paper.

## II. RELATED WORKS

This section reviews recent literature about Kalman filtering, cost matrix, and storage and update of appearance features, which are highly related to the focus of this study.

### A. Kalman Filtering

It is well known that Kalman filtering is a two-phase process involving prediction and update. In MOT tasks, Kalman filtering is frequently utilized to predict the current state of a trajectory based on its past state, and to update the trajectory state if it successfully matches a detection box.

Most existing MOT solutions, such as [3], [4], [5], [6], utilize fixed measurement noise covariance in the update stage of their Kalman filters, assuming that detection quality remains the same across all measurements (i.e., detection boxes). In fact, detection quality is subject to various factors, including illumination, distance, occlusion, etc. As a result, in actual MOT applications, objects' detection quality differs from each other, and their measurement noise covariance can never be exactly the same. In other words, the assumption of identical detection quality does not hold, and usage of this assumption in Kalman filtering inevitably jeopardizes data association efficacy.

To tackle the above limitations, Du et al. [7] enhanced the traditional Kalman filter by introducing the NSAKF, which adjusts measurement noise level based on detection quality, aiming to refine data association accuracy. Liu et al. [8] further advanced this approach with the PAKF [8], enabling separate adjustments of process noise and measurement noise using Intersection over Union (IoU) and detection confidence, respectively. Note that if a trajectory encounters occlusions or suddenly leaves the frame, it undergoes continuous prediction without state update, and consequently, the prediction error inevitably accumulates as time elapses. Hence, the prediction result should be less trusted, and more reliance should be placed on detection. However, this important issue has not yet been considered in the above methods.

Aiming to address this challenge, an adaptive Kalman filter is designed for more precise data association in this study. We propose an adaptive factor, dependent on detection confidence and trajectory disappearance, to dynamically adjust the measurement noise. This dynamic adjustment mechanism significantly enhances tracking performance by assigning a higher weight to more reliable information during Kalman filtering.

### B. Cost Matrix

Among existing TBD-based MOT algorithms, it is a common practice to construct a cost matrix for evaluating the similarity between trajectory boxes and detection boxes. This cost matrix plays a crucial role in ensuring effective data association and achieving competitive tracking performance.

Within the scope of relevant literature, it has been noted that many MOT solutions (e.g. [3], [13], [14], [15]) predominantly utilize the motion information-based IoU as the cost matrix, due to its simplicity in computation. However, a cost matrix that solely relies on motion information cannot guarantee effective data association, especially when dealing with objects re-emerging from prolonged occlusions, potentially leading to severe ID Switches (IDSW).

To tackle the above shortcoming, some approaches have incorporated neural networks to extract appearance features of detected objects, and employed the cosine distance between these features to evaluate the appearance similarity between trajectory and detection boxes. For example, Aharon et al. [9] introduced BoT-SORT, which employs either the motion-based IoU or the appearance-based cosine distance as the cost matrix. This approach, however, has not made full use of available information as only one type of information is used at a time. Wojke et al. [5] and Du et al. [10] also employed appearance similarity as the cost matrix, given that the motion cost (Mahalanobis distance [16], [17]) between the trajectory and detection boxes is below a certain threshold. Besides, Stadler et al. [11] proposed directly combining the motion-based IoU and the appearance-based cosine distance in the cost matrix, with the weights for IoU and cosine distance determined experimentally.

It is critical to note that none of the above methods realizes adaptive weighting adjustment for the motion and appearance costs; hence, the weighting factors are not calibrated according to detection quality (i.e., localization accuracy and appearance clarity). Aiming to bridge this gap, this study introduces a cost matrix that adaptively adjusts the weighting factors between motion and appearance costs, by using the localization accuracy and appearance clarity of each detection box. This cost matrix design considers the effects of detection quality, thereby significantly improving the efficacy of data association.

### C. Storage and Update of Appearance Features

In MOT applications, a mechanism is usually required to store and update objects' appearance features, facilitating re-identification of objects for situations such as occlusion, direction change, and re-appearance. This mechanism also plays an important role in differentiating between objects and contributes to the robustness of MOT methodologies.

DeepSORT, introduced by Wojke et al. [5], employs a storage mechanism to archive the appearance features of trajectory boxes from the last 100 frames. For a given detection box in the current frame, its appearance feature is compared with the 100 stored features in terms of cosine distance. Then, the minimum cosine distance is employed as the cost matrix. Once a detection box matches a trajectory box, the appearance feature of this detection box is directly added to the storage, without considering its detection quality. This approach, however, makes data association vulnerable to low-quality detections.

Subsequent MOT strategies, such as [4], [9], [10], have adopted the EMA approach for updating the appearance features of trajectories. The EMA method utilizes a fixed weighting factor to combine the appearance feature of the newly matched detection box and that of the trajectory box from the previous frame. Compared to the strategy used in DeepSORT, the EMA approach does not fully incorporate the appearance feature of the newly matched detection, but only a portion of it through this fixed weighting factor, which enhances robustness of data association against low-quality detections. However, this enhancement is limited because the fixed weighting factor fuses the newly matched detection regardless of its detection quality.

To address the above shortcomings, Maggiolino et al. [12] incorporated detection confidence in their DA strategy, which dynamically adjusts the weighting factor used in the EMA strategy. By this means, the detection quality of each detection box is effectively taken into consideration, and appearance features with different confidence levels are assigned different weights during feature updates. However, this strategy solely depends on the overall detection confidence (which contains both localization and appearance information) for weighting adjustments, and overlooks the individual impacts of localization accuracy and appearance clarity on appearance features.

This study proposes an innovative strategy for updating appearance features of trajectories, which separately accounts for the effects of localization accuracy and appearance clarity of detections. Our proposed approach no longer employs the fixed fusion pattern and allows for adaptive adjustment of the EMA weighting factor, which leads to more precise data association.

### III. METHOD

In this paper, we propose an improved TBD-based multi-object tracker based on our previous tracking framework – LG-Track [1]. We have made substantial improvements in three aspects: adaptive covariance of measurement noise, adaptive cost matrix, and strong dynamic appearance. The overall architecture of this new tracker is demonstrated in Fig. 2.

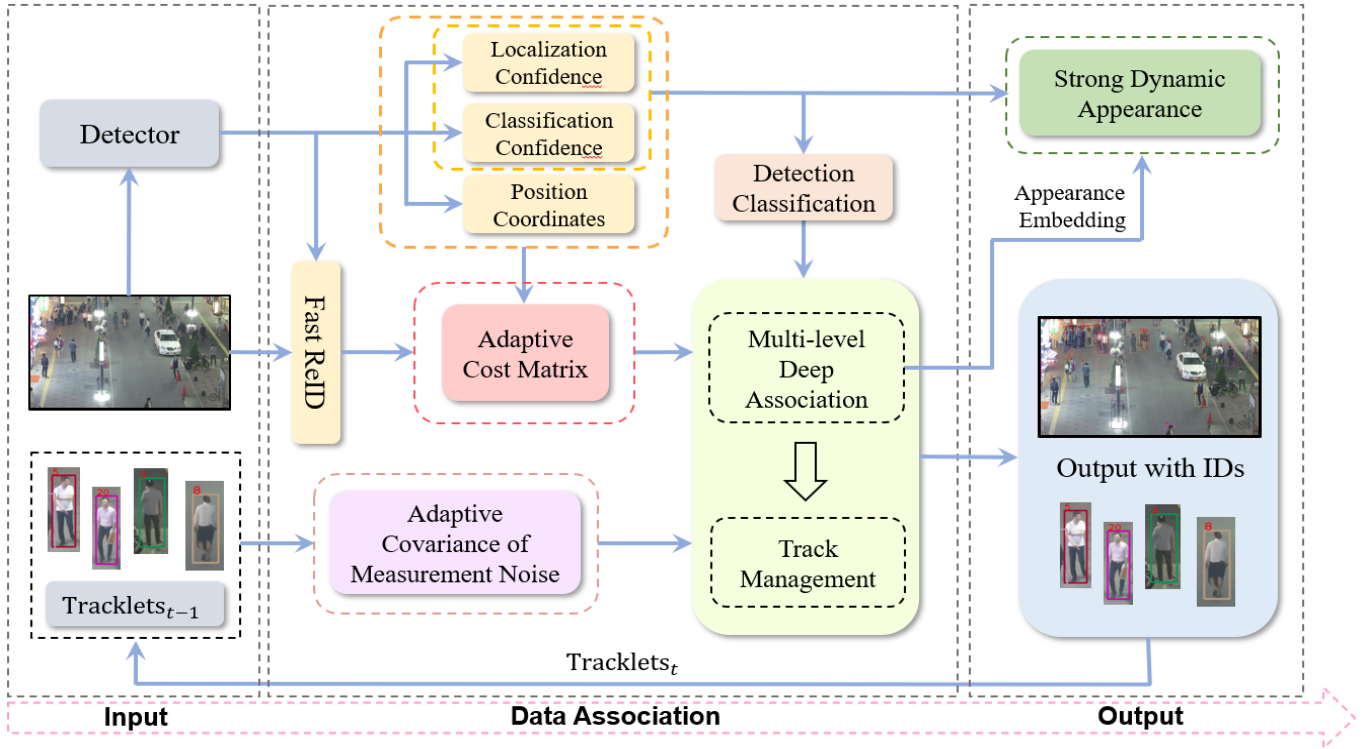

**Fig. 2.** Overall architecture of our proposed tracker.

*A. Adaptive Covariance of Measurement Noise*

In our previous tracking framework – LG-Track, a Kalman filter was used for each trajectory to predict its location in the new frame and then update its status. In each Kalman filter, the covariance matrix $R$ of measurement noise naturally reflects the level of uncertainty in the measurement. Intuitively, the higher this uncertainty is, the less reliable the measurement is.

In terms of the Kalman filter used in our previous work – LG-Track (and many other existing trackers, such as [1], [18], [19]), the level of measurement uncertainty is invariant. In other words, the covariance matrix $R$ of measurement noise is fixed. However, it must be noted that in practical MOT applications, uncertainty varies from measurement to measurement, even with exactly the same detector. Since measurement uncertainty naturally reflects the quality of measurement received, when measurement uncertainty is low, a higher weight should be assigned to this measurement (i.e. rely more on measurement than prediction) in the Kalman filtering process.

Considering the above facts, we introduce an adaptive covariance matrix to reflect the variation of measurement uncertainty. To do this, we categorize trajectories into four types of states, including "New", "Tracked", "Lost", and "Removed". Once a trajectory box is successfully matched with a detection box through data association, the detection confidence of this detection box is also affixed to the matched trajectory.

For "New" and "Tracked" trajectories, if they are matched with detections with high confidence (i.e., measurements are reliable), then we decrease the level of measurement uncertainty. Otherwise, we increase the level of measurement uncertainty. For "Lost" trajectories that are successfully re-matched after several frames of disappearance (e.g., occlusions), we decrease the level of measurement uncertainty to vary the relative weighting between prediction and measurement in the update process (i.e., increase reliance on measurement and reduce dependence on prediction). This is because these trajectories are propagated through mere predictions during disappearance. As a result, prediction error gradually accumulates, and predictions become increasingly unreliable.

Based on the above principles, we set a weighting factor $\alpha$ to adjust the level of measurement uncertainty for "New", "Tracked", and "Lost" trajectories:

$$\widetilde{R} = \alpha R \quad (1)$$

where $R$ denotes the preset covariance matrix of measurement noise, $\widetilde{R}$ indicates the proposed adaptive covariance matrix. The following formulation is proposed in this study to adaptively adjust the weighting factor $\alpha$:

$$\alpha = \begin{cases} \dfrac{th_{\text{det}}}{S_{\text{det}}}, & \text{if } S_{\text{det}} > th_{\text{det}} \\ (e^{1-S_{\text{det}}})^{1.5-N}, & \text{else} \end{cases} \quad (2)$$

$$N = \begin{cases} \dfrac{N_{\text{lost}}}{N_{\text{max}}}, & \text{if } \dfrac{N_{\text{lost}}}{N_{\text{max}}} > 0.5 \\ 0.5, & \text{else} \end{cases} \quad (3)$$

where $S_{\text{det}}$ represents the detection confidence affixed to the trajectory, $th_{\text{det}} = 0.6$ implies the detection confidence threshold, $N_{\text{lost}}$ indicates the number of consecutive frames that the lost trajectory has disappeared (for "New" and "Tracked" trajectories, $N_{\text{lost}} = 0$), and $N_{\max}$ signifies the maximum number of consecutive frames that a trajectory is retained after its disappearance.

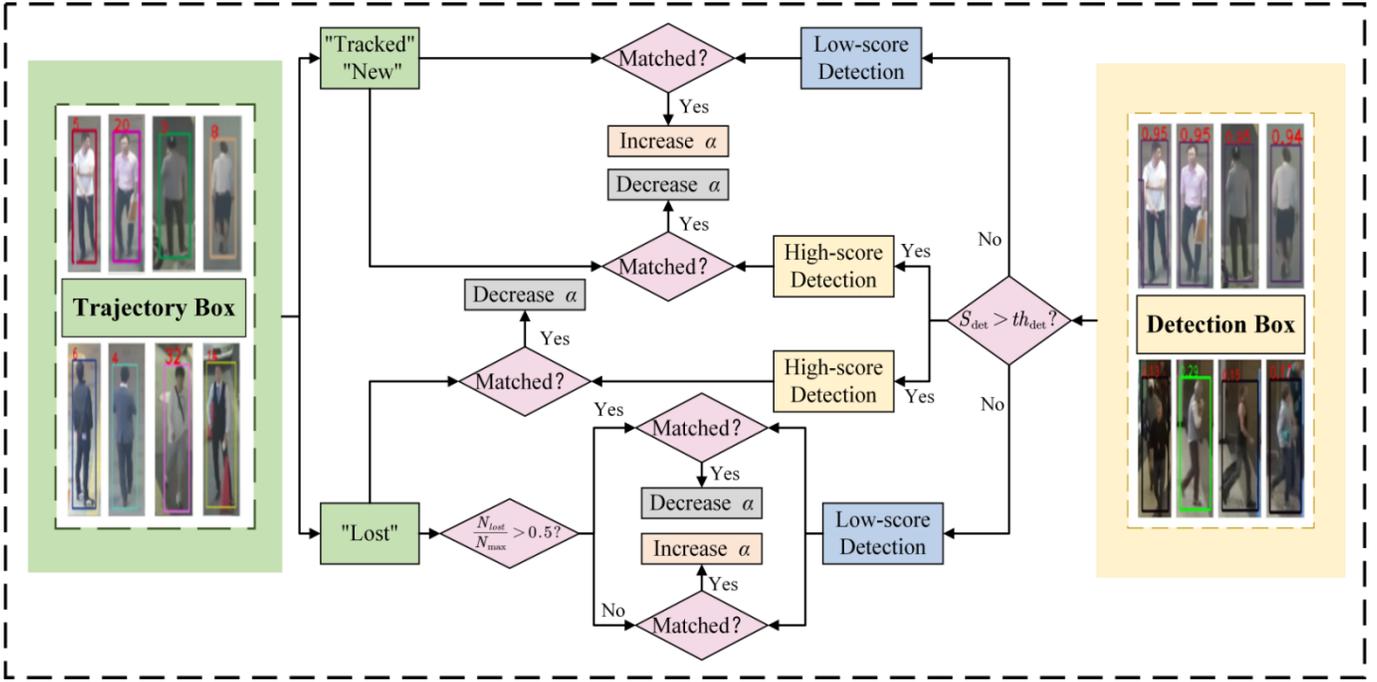

**Fig. 3.** Design process of the adaptive weighting factor $\alpha$.

As shown by equations (1), (2) and (3), the proposed covariance matrix $\widetilde{\boldsymbol{R}}$ is dependent on the quality of measurement received. Specifically, as the quality of measurement deteriorates (namely, the detection confidence $S_{\text{det}}$ decreases), all elements of the proposed covariance matrix $\widetilde{\boldsymbol{R}}$ are increased according to the pattern designed in equation (2), meaning that lower weighting is assigned to the measurement in the Kalman filtering process.

On the other hand, if a "Lost" trajectory is matched with a detection box with low confidence, the level of measurement uncertainty (i.e., the coefficient $\left(e^{1-S_{\text{det}}}\right)^{1.5-N}$ given in equation (2)) is decreased as the number of disappeared frames $N_{\text{lost}}$ increases. This means that higher weighting is assigned to the measurement in the Kalman filtering process, whilst the prediction result is less trusted due to accumulative prediction error.

The above process for designing the adaptive weighting factor $\alpha$ is schematically shown in Fig. 3.

*B. Adaptive Cost Matrix*

It has been pointed out in [4], [9], [10], [11] that utilizing both motion and appearance information in a cost matrix is beneficial to enhancing tracking performance, especially when dealing with complex scenarios such as dense crowds. However, existing trackers of such type present various limitations.

However, existing trackers of such type present various limitations. For example, Aharon et al. [9] proposed BoT-SORT, in which the cost matrix is designed as the smaller of the two – IoU and appearance cost. In this design, only one type of information (either motion or appearance) is used at a time, and no proper fusion mechanism is in place. Some other solutions in the literature, such as [4], [10], [11] employ a fixed weighting factor for fusing motion and appearance information. However, in real MOT applications, different "trajectory–detection" pairs usually present different motion and appearance characteristics. As an example, Fig. 4 demonstrates detection boxes in the MOT20 dataset with different motion and appearance characteristics. Note that above each detection box, the left number represents the localization confidence, while the right number denotes the classification confidence. In Fig. 4(a), both localization and classification confidences are high, indicating precise localization and clear appearance. In this case, both motion and appearance information can be reliably used for data association. In Fig. 4(b), the localization confidence scores are low (0.06 and 0.04), while the classification confidence scores remain high, suggesting that appearance similarity should be prioritized in the cost matrix. In Fig. 4(c), the localization confidence scores are high, but the classification confidence scores are relatively low due to occlusion. In these detection boxes, the detected pedestrians are occluded by other people, reducing the reliability of appearance information. From the above illustration, it is evident that using a fixed weighting factor can hardly produce robust matching performance.

Aiming to overcome the above shortcomings, in this study, we propose an adaptive cost matrix to achieve more accurate and robust data association.

Specifically, we take into consideration both localization confidence and detection confidence of detections, and employ them as weighting factors in the cost matrix for fusing motion information (IoU) and appearance information (appearance cost). The proposed adaptive cost matrix is written as follows:

$$C(i,j) = C_{\text{IoU}}(i,j) \times S_{\text{loc}}(i) + C_{\cos}(i,j) \times S_{\det}(i) \quad (4)$$

where $C(i,j)$ represents the cost between the $i$th detection box and the $j$th trajectory box, $C_{\text{IoU}}(i,j)$ denotes the IoU between these two boxes, $C_{\cos}(i,j)$ indicates the appearance similarity (measured by cosine distance) between these two boxes, $S_{\text{loc}}(i)$ refers to the localization confidence of the $i$th detection box, and $S_{\det}(i)$ stands for the detection confidence of the $i$th detection box. Note that in our previous tracking framework – LG-Track [1], different cost matrix was used in different level of data association. Unlike LG-Track, in this study, the proposed adaptive cost matrix – equation (4) – is uniformly used for all association levels.

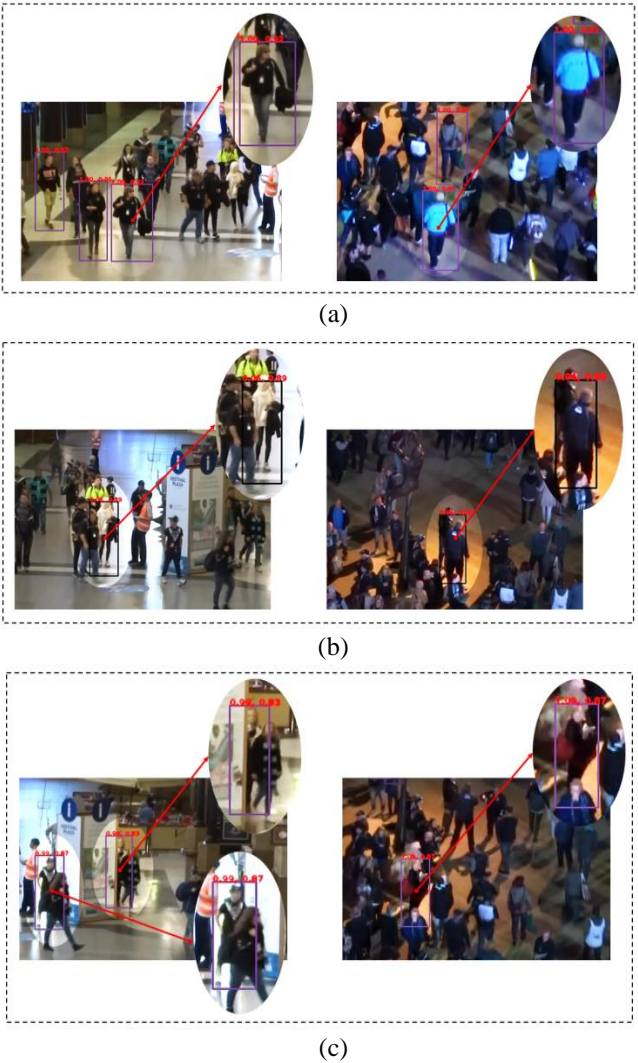

**Fig. 4.** Detection boxes in the MOT20 dataset with different motion and appearance characteristics.

In the above cost matrix design, the relative emphasis on motion information and appearance information is made adaptive to individual detections, by using localization and detection confidence as weighting factors. In other words, with the proposed cost matrix, more emphasis is given to information with higher reliability. For instance, if a detection box is obtained with high localization confidence and low detection confidence, then the relative weighting of IoU is increased, and that of appearance cost is reduced accordingly. This adaptive cost matrix design effectively enhances the accuracy and robustness of data association.

It must be pointed out that in equation (4), the weighting factor for appearance similarity (i.e. $C_{\cos}(i,j)$) is designed as the detection confidence instead of the classification confidence. The underlying reason is that the accuracy of appearance information depends not only on classification precision, but also on localization accuracy. Hence, it makes a lot more sense to employ the detection confidence, an index that integrates both classification confidence and localization confidence [20], [21], [22], [23], [24], as the weighting factor for appearance similarity.

### C. Strong Dynamic Appearance

As introduced in Section I, existing works [4], [9], [10] commonly relied on the EMA approach to update the appearance features of trajectory boxes, using a fixed factor to adjust the relative weighting between previous and current features. On this basis, in Deep OC-SORT [12], Maggiolino et al. proposed to employ the DA approach to dynamically tune this weighting factor using the detection confidence of detection boxes. However, we must note that the accuracy of appearance features depends not only on the level of appearance clarity, but also on the localization accuracy of detection boxes. As a result, tuning the above-mentioned weighting factor based simply on the detection confidence of detection boxes cannot ensure optimal tracking performance.

It has been pointed out in LG-Track [1] that for commonly used one-stage detectors (such as the YOLO series [20], [21], [22], [23], [24]), the detection confidence is produced to reflect not only classification accuracy (classification confidence) but also localization precision (localization confidence). Inspired by these indexes, we propose a novel feature updating mechanism – strong dynamic appearance (SDA), considering both appearance clarity and localization accuracy. Results in Section IV proved that our proposed SDA mechanism enhances data association efficacy, which in turn improves tracking performance.

The original EMA mechanism can be mathematically expressed as follows [4], [9], [10]:

$$\boldsymbol{e}_t = \alpha_{\text{EMA}} \boldsymbol{e}_{t-1} + (1 - \alpha_{\text{EMA}}) \boldsymbol{f} \quad (5)$$

where vectors $\boldsymbol{e}_t$ and $\boldsymbol{e}_{t-1}$ denote the appearance features of trajectory boxes in frames $t$ and $t-1$, vector $\boldsymbol{f}$ represents the appearance feature of the newly matched detection box, and $\alpha_{\text{EMA}}$ stands for a fixed weighting factor between vectors $\boldsymbol{e}_{t-1}$ and $\boldsymbol{f}$.

In Deep OC-SORT, the fixed weighting factor $\alpha_{\text{EMA}}$ used in the above EMA mechanism is replaced by the following

TABLE I
PERFORMANCE COMPARISON OF LEADING TRACKING METHODS ON THE MOT17 TEST SET. THE METHODS WITH LIGHT BLUE SHADING EMPLOY YOLOX [24] AS THE 2D DETECTOR. DATA MARKED IN RED AND BOLD INDICATE THE BEST FOR RESPECTIVE METRICS, WHILE DATA MARKED IN BLUE AND BOLD SIGNIFY THE SECOND BEST FOR RESPECTIVE METRICS.

| Method | Published Year | HOTA (%↑) | AssA (%↑) | IDF1 (%↑) | MOTA (%↑) | DetA (%↑) | IDSW (↓) |
|---|---|---|---|---|---|---|---|
| RTU++ [30] | TIP (2022) | 63.9 | 63.7 | 79.1 | 79.5 | 64.5 | 1302 |
| TrackFormer [31] | CVPR (2022) | 57.3 | 54.1 | 68.0 | 74.1 | 60.9 | 2829 |
| UTM [32] | CVPR (2023) | 64.0 | 62.5 | 78.7 | **81.8** | **65.9** | 1431 |
| BPMTrack [33] | TIP (2024) | 63.6 | 62.0 | 78.1 | 81.3 | 65.5 | 2010 |
| AttentionTrack [34] | TITS (2024) | ---- | ---- | 66.0 | 70.7 | 56.5 | 4722 |
| ByteTrack [15] | ECCV (2022) | 63.1 | 62.0 | 77.3 | 80.3 | 64.5 | 2196 |
| BoT-SORT [9] | ArXiv (2022) | 65.0 | 65.5 | 80.2 | 80.5 | 64.9 | 1212 |
| SAT [35] | MM (2023) | 64.4 | 64.4 | 79.8 | 80.0 | 64.8 | 1356 |
| GHOST [36] | CVPR (2023) | 62.8 | ---- | 77.1 | 78.7 | ---- | 2325 |
| StrongSORT++ [10] | TMM (2023) | 64.4 | 64.4 | 79.5 | 79.6 | 64.6 | 1194 |
| OC-SORT [14] | CVPR (2023) | 63.2 | 63.2 | 77.5 | 78.0 | ---- | 1950 |
| FineTrack [37] | CVPR (2023) | 64.3 | 64.5 | 79.5 | 80.0 | 64.5 | 1272 |
| MotionTrack [38] | CVPR (2023) | 65.1 | 65.1 | 80.1 | 81.1 | 65.4 | 1140 |
| Deep OC-SORT [12] | ICIP (2023) | 64.9 | 65.9 | 80.6 | 79.4 | ---- | **1023** |
| Hybrid-SORT-ReID [39] | AAAI (2024) | 64.0 | 63.5 | 78.7 | 79.9 | ---- | 1191 |
| ConfTrack [40] | WACV (2024) | 65.4 | **66.3** | **81.2** | 80.0 | 64.8 | 1677 |
| UCMCTrack [41] | AAAI (2024) | **65.7** | **66.4** | 81.0 | 80.6 | 65.3 | 1689 |
| PCL_YOLOX [42] | CVPR (2024) | 65.0 | 64.9 | 79.6 | 80.9 | 65.4 | 1749 |
| BUSCA [43] | ECCV (2024) | 63.9 | 64.2 | 79.2 | 78.6 | 63.9 | 1428 |
| SparseTrack [44] | TCSVT (2024) | 65.1 | 65.1 | 80.1 | 81.0 | 65.3 | 1170 |
| LG-Track [1] | IEEE SENS. J. (2025) | 65.4 | 65.4 | 80.4 | **81.4** | **65.6** | 1125 |
| **Deep LG-Track** | ---- | **65.9** | **66.4** | **81.4** | 81.3 | 65.5 | **999** |

detection-dependent factor $\alpha_{\text{DOC}}$ [12]:

$$\alpha_{\text{DOC}} = c + (1-c)\left(1 - \frac{S_{\text{det}} - th_{\text{det}}}{1 - th_{\text{det}}}\right) \quad (6)$$

where $S_{\text{det}}$ denotes the detection confidence, $th_{\text{det}} = 0.6$ implies the detection confidence threshold, and $c = 0.95$ is a user-defined constant.

As previously elaborated, the above weighting factor $\alpha_{\text{DOC}}$ is dependent merely on the detection confidence of detection boxes, and it cannot ensure optimal tracking performance. To overcome this shortcoming, in this study, we take into account not only appearance clarity but also localization accuracy, and propose the following SDA feature updating mechanism:

$$\begin{aligned}\alpha_{\text{SDA}} = c &+ c_{\text{cls}}\left(1 - \frac{S_{\text{cls}} - th_{\text{cls}}}{1 - th_{\text{cls}}}\right) \\ &+ c_{\text{loc}}\left(1 - \frac{S_{\text{loc}} - th_{\text{loc}}}{1 - th_{\text{loc}}}\right)\end{aligned} \quad (7)$$

where $S_{\text{cls}}$ represents the classification confidence, $S_{\text{loc}}$ denotes the localization confidence, $th_{\text{cls}} = 0.75$ implies the classification confidence threshold, $th_{\text{loc}} = 0.55$ indicates the localization confidence threshold, $c_{\text{cls}} = 4(1-c)/5$ and $c_{\text{loc}} = (1-c)/5$

By using our proposed SDA feature updating mechanism, the relative weighting between historical and current appearance features is made dependent on both appearance clarity and localization accuracy, which enhances data association performance.

## IV. EXPERIMENTS

In this section, both quantitative and qualitative results are presented to demonstrate the effectiveness of the proposed Deep LG-Track in comparison with state-of-the-art (SOTA) methods in the literature. Besides, ablation study results are also provided to exhibit the contribution of each constituent module within Deep LG-track. As previously mentioned, the three key improvements introduced in this paper have been integrated into our prior LG-Track framework. Building upon this foundation, the experiment results presented in the following subsections demonstrate the favorable impacts of these enhancements.

### A. Datasets and Evaluation Metrics

**Datasets:** We evaluated our proposed Deep LG-track against SOTA tracking methods on the commonly used MOT17 [25] and MOT20 [26] test sets under the "private detector" protocol. The results shown in Tables I and II were sourced from the leaderboards of MOT17 and MOT20 datasets. For consistency, we employed the publicly accessible YOLOX [24] detector trained by ByteTrack [15]. Object appearance data was extracted using the FastReID [27] SBS-50 model from BoT-SORT [9].

**Metrics:** In this study, two commonly used evaluation metrics, HOTA [28] and CLEAR [29], have been employed to investigate the performance of competing trackers. These metrics can be broken down into several essential sub-metrics, such as AssA, DetA, MOTA, IDSW, and IDF1.

TABLE II

PERFORMANCE COMPARISON OF LEADING TRACKING METHODS ON THE MOT20 TEST SET. THE METHODS WITH LIGHT BLUE SHADING EMPLOY YOLOX [24] AS THE 2D DETECTOR. DATA MARKED IN RED AND BOLD INDICATE THE BEST FOR RESPECTIVE METRICS, WHILE DATA MARKED IN BLUE AND BOLD SIGNIFY THE SECOND BEST FOR RESPECTIVE METRICS.

| Method | Published Year | HOTA (%↑) | AssA (%↑) | IDF1 (%↑) | MOTA (%↑) | DetA (%↑) | IDSW (↓) |
|---|---|---|---|---|---|---|---|
| RTU++ [30] | TIP (2022) | 62.8 | 62.6 | 76.8 | 76.5 | 63.1 | 971 |
| TrackFormer [31] | CVPR (2022) | 54.7 | 53.0 | 65.7 | 68.6 | 56.7 | 1532 |
| UTM [32] | CVPR (2023) | 62.5 | 61.4 | 76.9 | **78.2** | 63.7 | 1228 |
| BPMTrack [33] | TIP (2024) | 62.3 | 60.9 | 76.7 | **78.3** | 63.9 | 1314 |
| TSAEN [45] | TITS (2022) | ---- | ---- | 64.3 | 62.0 | ---- | 1182 |
| ByteTrack [15] | ECCV (2022) | 61.3 | 59.6 | 75.2 | 77.8 | 63.4 | 1223 |
| BoT-SORT [9] | ArXiv (2022) | 63.3 | 62.9 | 77.5 | 77.8 | **64.0** | 1313 |
| SAT [35] | MM (2023) | 62.6 | 63.2 | 76.6 | 75.0 | 62.1 | 816 |
| GHOST [36] | CVPR (2023) | 61.2 | ---- | 75.2 | 73.7 | ---- | 1264 |
| StrongSORT++ [10] | TMM (2023) | 62.6 | 64.0 | 77.0 | 73.8 | 61.3 | **770** |
| OC-SORT [14] | CVPR (2023) | 62.1 | 62.0 | 75.9 | 75.5 | ---- | 913 |
| FineTrack [37] | CVPR (2023) | 63.6 | 63.8 | **79.0** | 77.9 | 63.6 | 980 |
| MotionTrack [38] | CVPR (2023) | 62.8 | 61.8 | 76.5 | 78.0 | **64.0** | 1165 |
| Deep OC-SORT [12] | ICIP (2023) | **63.9** | **65.7** | **79.2** | 75.6 | ---- | **779** |
| Hybrid-SORT-ReID [39] | AAAI (2024) | **63.9** | 64.5 | 78.4 | 76.7 | ---- | 1136 |
| UCMCTrack [41] | AAAI (2024) | 62.8 | 63.5 | 77.4 | 75.6 | 62.4 | 1335 |
| BUSCA [43] | ECCV (2024) | 61.8 | 63.5 | 76.3 | 72.7 | 60.3 | 1006 |
| SparseTrack [44] | TCSVT (2024) | 63.4 | ---- | 77.3 | **78.2** | ---- | 1116 |
| LG-Track [1] | IEEE SENS. J. (2025) | 63.4 | 62.9 | 77.4 | 77.8 | **64.1** | 1161 |
| **Deep LG-Track** | ---- | **64.0** | 64.1 | 78.4 | 77.6 | **64.0** | 1081 |

*B. Comparison with the State-of-the-Art Trackers*

**Quantitative results:** We tested our proposed Deep LG-Track in comparison with SOTA trackers on the MOT17 and MOT20 test sets. The quantitative results are demonstrated in Tables I and II.

As shown in Table I, on the MOT17 test set, our proposed Deep LG-Track achieved the best results in terms of HOTA (65.9), AssA (66.4), IDF1 (81.4), and IDSW (999). This indicates that on the MOT17 test set, Deep LG-Track produced the best overall tracking performance among all SOTA trackers compared. Besides, we see in Table II that on the MOT20 test set, the proposed Deep LG-Track attained the highest HOTA (64.0), the second highest DetA (64.0), the third highest AssA (64.1), and the third highest IDF1 (78.4). This reflects that Deep LG-Track also presents impressive tracking performance on the MOT20 test set.

Note that compared to the original LG-Track, our newly proposed Deep LG-Track significantly improved the AssA metric. Specifically, for the MOT17 test set, the AssA was improved by 1.0, while for the MOT20 test set, the AssA was increased by 1.2. Since the AssA metric is an index that reflects the efficacy of data association, the above quantitative results demonstrate that the three important improvements we proposed in this paper effectively enhanced the data association performance of our previous work – LG-Track.

**Qualitative results:** Fig. 5 and Fig. 6 demonstrate the visualization results of the proposed Deep LG-Track in comparison with our previous LG-Track, on the MOT17-05 and MOT17-10 sequences. As for LG-Track, by comparing frame 296 and frame 302 in Fig. 5, we clearly see in frame 296 that the man wearing a black T-shirt was assigned ID 56 and leaving the scene, who was no longer present in frame 302. However, in frame 302, ID 56 was assigned to a different person wearing a white T-shirt who entered the scene from the left of the image. Apparently, these two pedestrians were mistaken for each other by LG-Track. Furthermore, in frame 310, we observe that the ID of the man wearing a white T-shirt was changed from 56 to 66, indicating an occurrence of ID switch. Also, the lady with ID 2 in frame 296 and frame 302 was altered to ID 36. From the visualization results above, we can see that in the MOT17-05 sequence, our previous LG-Track encountered challenges in the tracking process.

In contrast, the newly proposed Deep LG-Track exhibited significantly more robust and reliable tracking performance. Specifically, Deep LG-Track assigned a new ID to the man wearing a white T-shirt, who newly entered the scene in frame 302. Namely, with Deep LG-Track, this man was no longer mistaken for the man wearing a black T-shirt, and stable tracking was achieved without incurring any ID switches in the following frames. The superiority of Deep LG-Track is also observed in Fig. 6. We see that on the MOT17-10 sequence, in frames 276, 288, and 360, the ID of the same pedestrian with a black backpack was 59, 48, and 64, respectively, indicating two ID switches when tracked by our precious LG-Track. In comparison, Deep LG-Track stably tracked the same person without incurring any ID switches, showing significantly better robustness than LG-Track.

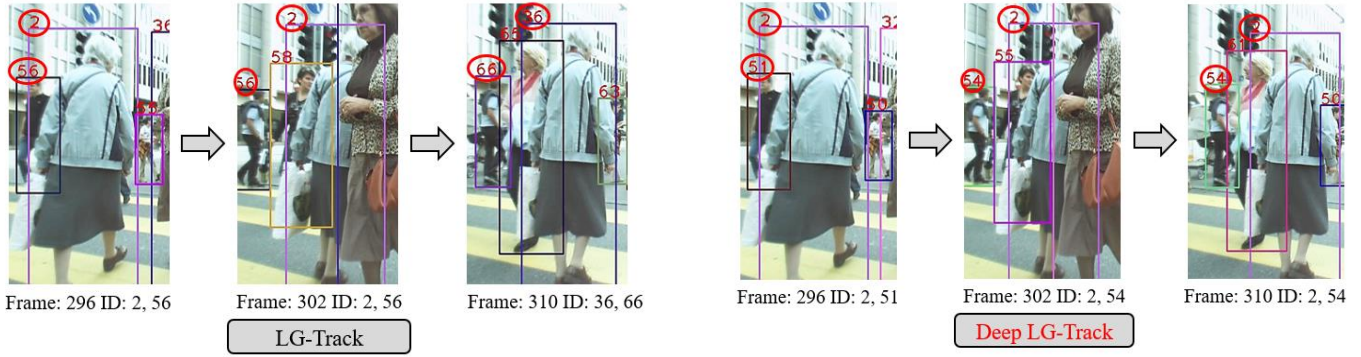

**Fig. 5.** Comparison of the proposed Deep LG-Track with LG-Track in terms of visualized results on MOT17-05.

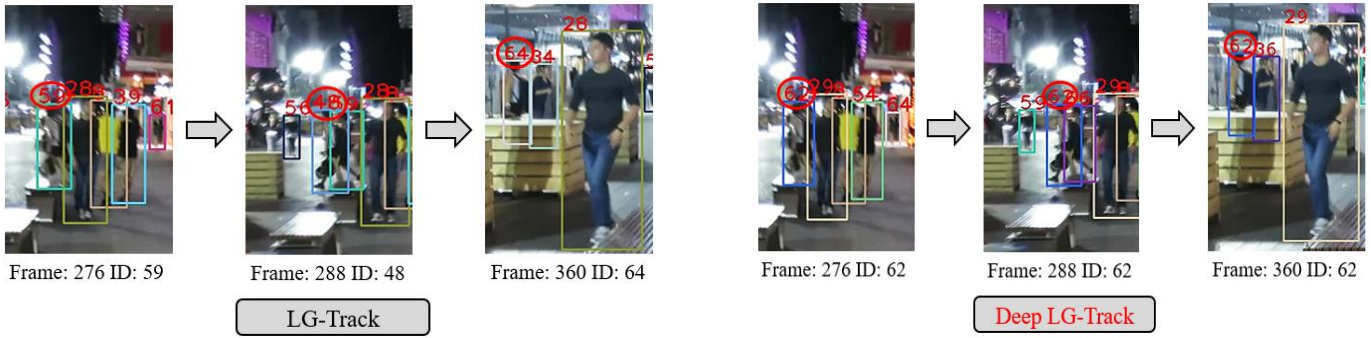

**Fig. 6.** Comparison of the proposed Deep LG-Track with LG-Track in terms of visualized results on MOT17-10.

*C. Ablation Study*

To reveal the effects of the three improvements proposed in Section III, we conducted a thorough ablation study on the MOT17 training set. Using our previous LG-Track as the baseline, the three improvements are added to LG-Track, one at a time. The results of this ablation study are given in Table III.

We see from the results in Table III that all three improvements proposed in this paper are effective and significantly boost the performance of the original LG-Track. Specifically, it is shown that the first proposed improvement – adaptive covariance of measurement noise (ACMN) – increased all four metrics, leading to enhanced overall tracking performance. Besides, the other two proposed improvements, i.e., adaptive cost matrix (ACM) and strong dynamic appearance (SDA), both enhanced tracking performance by elevating HOTA, AssA, and IDF1. Note that among the above three improvements, the ACM contributed to the largest increase of HOTA (i.e., 0.54%) – the most important performance metric.

Besides, it has been mentioned in the Introduction that the NSAKF [5] was devised in GIAOTracker to handle the measurement noise. To demonstrate the advantage of our proposed ACMN over NSAKF, in this ablation study we compared NSAKF and ACMN within three typical tracking frameworks – ByteTrack, BoT-SORT, and LG-Track. The comparison results are given in Table IV. It is clearly seen that compared with NSAKF, the proposed ACMN enhanced the performances of all three trackers, in terms of all four performance metrics.

Moreover, as we mentioned previously, the DA strategy was proposed in Deep OC-SORT to dynamically tune the weighting factor between previous and current appearance features. To reveal the advantage of our proposed SDA over the original DA, we compared the effects of DA and SDA on tracking performance using the MOT17 training set. We observe from Table V that with our SDA onboard, the tracking performance was enhanced in terms of HOTA, AssA, and IDF1. By introducing the SDA strategy, the AssA metric – which reflects data association efficacy – witnessed the largest enhancement among all metric.

## V. Conclusion

In this paper, we proposed Deep LG-Track as an enhanced version of our previously developed multi-object tracker, LG-Track, incorporating three key improvements. First, the Kalman filtering process has been optimized by adapting the covariance of measurement noise based on detection confidence and trajectory disappearance. This adaptation has been demonstrated to be effective when integrated into various SOTA trackers. Second, localization confidence and detection confidence are introduced as weighting factors in the cost matrix to fuse motion and appearance information, leading to more accurate and robust data association. Third, a dynamic appearance feature updating mechanism is proposed, where the relative weighting between historical and current appearance features is adjusted based on both appearance clarity and localization accuracy. This mechanism significantly enhances data association performance. The effectiveness of Deep LG-Track has been validated on the

TABLE III

ABLATION STUDY ON THE THREE PROPOSED IMPROVEMENTS – ACMN, ACM, AND SDA – USING THE MOT17 TRAINING SET (BEST RESULTS PRINTED IN BOLD).

| ACMN | ACM | SDA | HOTA (%↑) | AssA (%↑) | IDF1 (%↑) | MOTA (%↑) |
|---|---|---|---|---|---|---|
| -- | -- | -- | 77.72 | 76.35 | 86.54 | 90.68 |
| ✓ | ✗ | ✗ | 78.06 (+0.34) | 77.01 (+0.66) | 87.04 (+0.5) | **90.69** (+0.01) |
| ✓ | ✓ | ✗ | 78.60 (+0.88) | 77.63 (+1.28) | 87.89 (+1.35) | 90.62 (−0.06) |
| ✓ | ✓ | ✓ | **78.81** (+1.09) | **78.12** (+1.77) | **88.12** (+1.58) | 90.43 (−0.25) |

TABLE IV

COMPARISON BETWEEN NSAKF AND ACMN ON THE MOT17 TRAINING SET UNDER VARIOUS TRACKING FRAMEWORKS (BEST RESULTS PRINTED IN BOLD).

| Tracker | NSAKF | ACMN | HOTA (%↑) | AssA (%↑) | IDF1 (%↑) | MOTA (%↑) |
|---|---|---|---|---|---|---|
| **ByteTrack** | ✗ | ✗ | 74.76 | 71.98 | 83.10 | 90.02 |
|  | ✓ | ✗ | **75.80** (+1.04) | 72.99 (+1.01) | 83.74 (+0.64) | 90.17 (+0.15) |
|  | ✗ | ✓ | 75.73 (+0.97) | **73.31** (+1.33) | **84.21** (+1.11) | **90.29** (+0.27) |
| **BoT-SORT** | ✗ | ✗ | 76.75 | 74.20 | 85.28 | 90.97 |
|  | ✓ | ✗ | 76.44 (−0.31) | 73.79 (−0.41) | 84.63 (−0.65) | 90.84 (−0.13) |
|  | ✗ | ✓ | **77.13** (+0.38) | **74.80** (+0.60) | **85.99** (+0.71) | **91.00** (+0.03) |
| **LG-Track** | ✗ | ✗ | 77.72 | 76.35 | 86.54 | 90.68 |
|  | ✓ | ✗ | 77.39 (−0.33) | 75.79 (−0.56) | 86.20 (−0.34) | 90.59 (−0.09) |
|  | ✗ | ✓ | **78.06** (+0.34) | **77.01** (+0.66) | **87.04** (+0.50) | **90.69** (+0.01) |

TABLE V

COMPARISON BETWEEN DA AND SDA ON THE MOT17 TRAINING SET. (BEST RESULTS PRINTED IN BOLD)

|  | HOTA (%↑) | AssA (%↑) | IDF1 (%↑) | MOTA (%↑) |
|---|---|---|---|---|
| LG-Track+ACMN+ ACM | 78.60 | 77.63 | 87.89 | 90.62 |
| LG-Track+ ACMN + ACM +DA | 78.60 (--) | 77.54 (−0.09) | 87.72 (−0.17) | 90.47 (−0.15) |
| LG-Track+ ACMN + ACM +SDA | **78.81** (+0.21) | **78.12** (+0.49) | **88.12** (+0.23) | 90.43 (−0.19) |

MOT17 and MOT20 datasets, where it exhibits distinct advantages over multiple SOTA trackers across various performance metrics.